\documentclass[runningheads,a4paper]{llncs}

\usepackage{amssymb}
\usepackage{graphicx}

\usepackage{url}
\urldef{\mailsa}\path|{alfred.hofmann, ursula.barth, ingrid.haas, frank.holzwarth,|
\urldef{\mailsb}\path|anna.kramer, leonie.kunz, christine.reiss, nicole.sator,|
\urldef{\mailsc}\path|erika.siebert-cole, peter.strasser, lncs}@springer.com|    
\newcommand{\keywords}[1]{\par\addvspace\baselineskip
\noindent\keywordname\enspace\ignorespaces#1}

\usepackage{times}
\usepackage{graphicx} 
\usepackage{subfigure} 

\usepackage{mydefs}

\usepackage{hyperref}

\usepackage{amsfonts,mathrsfs,amsmath}
\usepackage{times}
\usepackage{graphicx} 

\usepackage{wasysym,wrapfig,subfigure,epsfig,shapepar,array,colortbl,bm,url}

\definecolor{DarkGreen}   {rgb}{0,0.5,0}
\definecolor{DarkBlue}    {rgb}{0,0.0,0.5}
\definecolor{LightGray}   {rgb}{0.8,0.8,0.8}

\newcommand{\beq}{\vspace{0mm}\begin{equation}}
\newcommand{\eeq}{\vspace{0mm}\end{equation}}
\newcommand{\beqs}{\vspace{0mm}\begin{eqnarray}}
\newcommand{\eeqs}{\vspace{0mm}\end{eqnarray}}
\newcommand{\barr}{\begin{array}}
      \newcommand{\earr}{\end{array}}

\newcommand{\Umat}[0]{{{\bf U}}}

\newcommand{\iv}{\boldsymbol{i}}

\newcommand{\uv}{\boldsymbol{u}}

\newcommand{\xv}{\boldsymbol{x}}

\newcommand{\zv}{\boldsymbol{z}}

\newcommand{\lambdav}{\boldsymbol{\lambda}}

\newcommand{\Ycal}{\mathcal{Y}}

\begin{document}

\mainmatter  

\title{Scalable Bayesian Non-Negative Tensor Factorization for Massive Count Data}

\titlerunning{Scalable Bayesian Non-Negative Tensor Factorization for Massive Count Data}

%
%
 \author{Changwei Hu$^*$
 \and Piyush Rai$^*$\and Changyou Chen$^*$\and \\ Matthew Harding$^\dagger$\and
 Lawrence Carin$^*$}

\authorrunning{Hu, Rai, Chen, Harding, Carin}

 \institute{$^*$ Department of Electrical \& Computer Engineering, Duke University\\
\texttt{\{ch237,piyush.rai,cc448,lcarin\}@duke.edu}\\
$^\dagger$Sanford School of Public Policy \& Department of Economics, Duke University\\
\texttt{matthew.harding@duke.edu}
}

%
%

\toctitle{Lecture Notes in Computer Science}
\tocauthor{Authors' Instructions}
\maketitle
\vspace{-2em}
\begin{abstract}
We present a Bayesian non-negative tensor factorization model for count-valued tensor data, and develop scalable inference algorithms (both batch and online) for dealing with massive tensors. Our generative model can handle overdispersed counts as well as infer the rank of the decomposition. Moreover, leveraging a reparameterization of the Poisson distribution as a multinomial facilitates conjugacy in the model and enables simple and efficient Gibbs sampling and variational Bayes (VB) inference updates, with a computational cost that only depends on the number of nonzeros in the tensor. The model also provides a nice interpretability for the factors; in our model, each factor corresponds to a ``topic''. We develop a set of online inference algorithms that allow further scaling up the model to massive tensors, for which batch inference methods may be infeasible. We apply our framework on diverse real-world applications, such as \emph{multiway} topic modeling on a scientific publications database, analyzing a political science data set, and analyzing a massive household transactions data set.
\vspace{-0.5em}
\end{abstract}
\vspace{-3em}
      \section{Introduction}
      \vspace{-0.5em}
      Discovering interpretable latent structures in complex multiway (tensor) data is an important problem when learning from polyadic relationships among multiple sets of objects. 
      Tensor factorization~\cite{kolda2009tensor,cichocki2009nonnegative} offers a promising way of extracting such latent structures. The inferred factors can be used to analyze objects in each mode of the tensor (e.g., via classification or clustering using the factors), or to do tensor completion.
      
      Of particular interest, in the context of such data, are sparsely-observed \emph{count-valued} tensors. Tensors are routinely encountered in many applications. For example, in analyzing a database of scientific publications, the data may be in form of a sparse four-way count-valued tensor (authors $\times$ words $\times$ journals $\times$ years). Another application where multiway count data is routinely encountered is the analysis of contingency tables~\cite{johndrow2014tensor} which represent the co-occurrence statistics of multiple sets of objects.
      
      We present a scalable Bayesian model for analyzing such sparsely-observed tensor data. Our framework is based on a beta-negative binomial construction, which provides a principled generative model for tensors with sparse and potentially overdispersed count data, and produces a non-negative tensor factorization. In addition to performing non-negative tensor factorization and tensor completion for count-valued tensors, our model has the property that each latent factor inferred for a tensor mode also represents a \emph{distribution} (or ``topic'', as in topic models) over the objects of that tensor mode; our model naturally accomplishes this by placing a Dirichlet prior over the columns of the factor matrix of each tensor mode. In addition to providing an expressive and interpretable model for analyzing sparse count-valued tensors, the model automatically infers the rank of the decomposition, which side-steps the crucial issue of pre-specifying the rank of the decomposition~\cite{kolda2009tensor,rai14tensor,zhao2014bayesian}. 
      
      Our framework also consists of a set of batch and scalable online inference methods. Using a reparameterization of the Poisson distribution as a multinomial allows us to achieve conjugacy, which facilitates closed-form Gibbs sampling as well as variational Bayes (VB) inference. Moreover, we also develop two \emph{online} inference algorithms - one based on online MCMC~\cite{guhaniyogi2014bayesian} and the other based on stochastic variational inference~\cite{hoffman2013stochastic}. These inference algorithms enable scaling up the model to massive-sized tensor data.
      
      One of the motivations behind our work is analyzing massive multiway data for tasks such as understanding thematic structures in scholarly databases (e.g., to design better recommender systems for scholars), understanding consumer behavior from shopping patterns of large demographies (e.g., to design better marketing and supply strategies), and understanding international relations in political science studies. In our experiments, we provide qualitative analyses for such applications on large-scale real-world data sets, and the scalability behavior of our model. 
      
			\vspace{-1em}
      \section{Canonical PARAFAC Decomposition}
			\vspace{-1em}
      \label{sec:cpdec}
      Given a tensor $\Ycal$ of size $n_1\times n_2 \times \cdots \times n_K$, with $n_{k}$ denoting the size of $\mathcal{Y}$ along the $k^{th}$ mode (or ``way'') of the tensor, the goal in a Canonical PARAFAC (CP) decomposition~\cite{kolda2009tensor} is to decompose $\Ycal$ into a set of $K$ factor matrices $\Umat^{(1)},\ldots,\Umat^{(K)}$ where $\Umat^{(k)} = [\uv_1^{(k)},\ldots,\uv_R^{(k)}], \ k = \{1,\ldots,K\}$, denotes the $n_k \times R$ factor matrix associated with mode $k$. In its most general form, CP decomposition expresses the tensor $\Ycal$ via a weighted sum of $R$ rank-1 tensors as $\Ycal \sim f(\sum_{r=1}^R \lambda_r . \uv_r^{(1)} \odot \ldots \odot \uv_r^{(K)})$.
      The form of $f$ depends on the type of data being modeled (e.g., $f$ can be Gaussian for real-valued, Bernoulli-logistic for binary-valued, Poisson for count-valued tensors). Here $\lambda_r$ is the weight associated with the $r^{th}$ rank-1 component, the $n_k\times 1$ column vector $\uv_r^{(k)}$ represents the $r^{th}$ latent factor of mode $k$, and $\odot$ denotes vector outer product. 
			\vspace{-0.5em}
      \section{Beta-Negative Binomial CP Decomposition}
			\vspace{-0.5em}
      \label{sec:bnbcp}
      
      We focus on modeling count-valued tensor data~\cite{chi2012tensors} and assume the following generative model for the tensor $\Ycal$
\vspace{-1em}
      \beqs
      \Ycal &\sim& \text{Pois}(\sum_{r=1}^R \lambda_r . \uv_r^{(1)} \odot \ldots \odot \uv_r^{(K)})\label{eq:poistensor} \\
      \uv^{(k)}_r &\sim& \text{Dir}(a^{(k)},\dots,a^{(k)})\label{eq:uprior}\\
      \lambda_r&\sim& \text{Gamma}(g_r,\frac{p_r}{1-p_r})\label{eq:lambdaprior}\\
      p_r&\sim& \text{Beta}(c\epsilon,c(1-\epsilon))\label{eq:prprior}
      \vspace{-0.25em}
      \eeqs 
      
      We use subscript $\iv = \{i_1,\ldots,i_K\}$ to denote the index of the $\iv$-th entry in $\Ycal$. Using this notation, the $\iv$-th entry of the tensor can be written as $y_{\vec{i}} \sim \text{Pois}(\sum_{r=1}^R \lambda_r \prod_{k=1}^K u_{i_k r}^{(k)})$. We assume that we are given $N$ observations $\{y_{\iv}\}_{\iv=1}^N$ from the tensor $\Ycal$.
       
      Since the gamma-Poisson mixture distribution is equivalent to a negative binomial distribution~\cite{kozubowski2008distributional}, (\ref{eq:poistensor}) and (\ref{eq:lambdaprior}), coupled with the beta prior (Eq~\ref{eq:prprior}) on $p_r$, lead to what we will call the beta-negative binomial CP (BNBCP) decomposition model. A few things worth noting about our model are
      \begin{itemize}
       \item The Dirichlet prior on the factors $\uv^{(k)}_r$ naturally imposes non-negativity constraints~\cite{chi2012tensors} on the factor matrices $\Umat^{(1)},\ldots,\Umat^{(K)}$. Moreover, since each column $\uv^{(k)}_r$ of these factor matrices sums to 1, $\uv^{(k)}_r$ can also be thought of a distribution (e.g., a ``topic'') over the $n_k$ entities in mode $k$. 
       \item The gamma-beta hierarchical construction of $\lambda_r$ (Eq~\ref{eq:lambdaprior} and ~\ref{eq:prprior}) allows inferring the rank of the tensor by setting an upper bound $R$ on the number of factors and letting the inference procedure infer the appropriate number of factors by shrinking the coefficients $\lambda_r$'s to close to zero for the irrelevant factors. 
       \item The resulting negative binomial model is useful for modeling \emph{overdispersed} count data in cases where the Poisson likelihood may not be suitable.
       \item Using alternate parameterizations (Section~\ref{sec:repar}) of the Poisson distribution in (\ref{eq:poistensor}) leads to a fully conjugate model and facilitates efficient Gibbs sampling and variational Bayes (VB) inference, in both batch as well as online settings.
      \end{itemize}
      
     
      \subsection{Reparametrizing the Poisson Distribution}
      \label{sec:repar}

      The generative model described in Eq (\ref{eq:poistensor})-(\ref{eq:prprior}) is not conjugate. We now describe two equivalent parametrizations~\cite{dunson2005bayesian,zhou2012NBPFA} of (\ref{eq:poistensor}), which transform (\ref{eq:poistensor})-(\ref{eq:prprior}) into a fully conjugate model and facilitate easy-to-derive and scalable inference procedures. These parameterizations are based on a data augmentation scheme described below.
      
      The first parametrization expresses the $\iv$-th count-valued entry $y_{\vec{i}}$ of the tensor $\Ycal$ as a sum of $R$ \emph{latent} counts $\{\tilde{y}_{\vec{i}r}\}_{r=1}^R$
			\vspace{-1em}
      \begin{equation}
      \label{eq:latentcounts}
      y_{\vec{i}} = \sum_{r=1}^R \tilde{y}_{\vec{i}r}, \quad \tilde{y}_{\vec{i}r}\sim \text{Pois}(\lambda_r \prod_{k=1}^K u_{i_k r}^{(k)})
      \end{equation}

      The second parametrization assumes the vector $\{\tilde{y}_{\vec{i}r}\}_{r=1}^R$ of latent counts is drawn from a multinomial as
      \beqs
      && \tilde{y}_{\vec{i}1},\ldots,\tilde{y}_{\vec{i}R} \sim \text{Mult}(y_{\vec{i}};\zeta_{\vec{i}1},\ldots,\zeta_{\vec{i}R}) \nonumber \\
      && \zeta_{\vec{i}r} = \frac{\lambda_r \prod_{k=1}^K u_{i_k r}^{(k)}}{\sum_{r=1}^R \lambda_r \prod_{k=1}^K u_{i_k r}^{(k)}} \label{eq:multsample}
      \eeqs
      
      The above parameterizations follows from the following lemma~\cite{dunson2005bayesian,zhou2012NBPFA}:
      \begin{lemma}
       Suppose that $x_1,\ldots,x_R$ are independent random variables with $x_r \sim \text{Pois}(\theta_r)$ and $x = \sum_{r=1}^R x_r$. Set $\theta = \sum_{r=1}^R\theta_r$; let $(z,z_1,\ldots,z_R)$ be another set of random variables such that $z \sim \text{Pois}(\theta)$, and $(z_1,\ldots,z_R)|z \sim \text{Mult}(z;\frac{\theta_1}{\theta},\ldots,\frac{\theta_R}{\theta})$. Then the distribution of $\xv = (x,x_1,\ldots,x_R)$ is the same as the distribution of $\zv = (z,z_1,\ldots,z_R)$.
      \end{lemma}
      
      These parameterizations, along with the fact that the columns $\uv_r^{(k)}$ of each factor matrix are drawn from a Dirichlet, allows us to leverage the Dirichlet-multinomial conjugacy and derive simple Gibbs sampling and variational Bayes (VB) inference update equations, as described in Section~\ref{sec:infer}.      
           \vspace{-1em}
	\section{Inference}
	\label{sec:infer}

	We first present the update equations for batch Gibbs sampling (Section~\ref{sec:batchgibbs}) and batch VB inference (Section~\ref{sec:batchvb}). We then present two online inference algorithms, based on: ($i$) conditional density filtering~\cite{guhaniyogi2014bayesian}, which provides an efficient way to perform online MCMC sampling using conditional sufficient statistics of the model parameters; and ($ii$) stochastic variational inference~\cite{hoffman2013stochastic}, which will allow scaling up VB inference by processing data in small minibatches.
	
	We also define two quantities $s_{j,r}^{(k)}=\sum_{\iv:i_k=j}\tilde{y}_{\iv r}$ and $s_r=\sum_{\iv}\tilde{y}_{\iv,r}$ which denote aggregates (sufficient statistics) computed using the latent counts $\tilde{y}_{\iv r}$. These quantities appear at various places in the description of the inference algorithms we develop.
	
	\subsection{Gibbs Sampling}
	\label{sec:batchgibbs}
	\begin{itemize}
	\item \textbf{Sampling $\tilde{y}_{\iv r}$:} The latent counts $\{\tilde{y}_{\vec{i}r}\}_{r=1}^R$ are sampled from a multinomial (\ref{eq:multsample}).\\
	\item \textbf{Sampling $\uv_r^{(k)}$:} Due to the Dirichlet-multinomial conjugacy, the columns of each factor matrix have Dirichlet posterior and are sampled as 
	\beq
	\uv_r^{(k)}\sim \text{Dir}(a^{(k)}+ s_{1,r}^{(k)},a^{(k)}+s_{2,r}^{(k)},\ldots,a^{(k)}+s_{n_k,r}^{(k)})
	\eeq	
	\item \textbf{Sampling $p_r$:}
	Using the fact that $s_r=\sum_{\iv}\tilde{y}_{\iv,r}$ and marginalizing over the $u_{i_k r}^{(k)}$'s in (\ref{eq:latentcounts}), we have $s_r \sim \text{Pois}(\lambda_r)$. Using this, along with (\ref{eq:lambdaprior}), we can express $s_r$ using a negative binomial distribution, i.e., $s_r\sim \text{NB}(g_r,p_r)$. Then, due to the conjugacy between negative binomial and beta, we can sample $p_r$ as 
	\beq
	p_r\sim \text{Beta}(c\epsilon+s_r,c(1-\epsilon)+g_r)
	\eeq 
	\item \textbf{Sampling  $\lambda_r$:} Again using the fact that $s_r \sim \text{Pois}(\lambda_r)$, and due to the gamma-Poisson conjugacy, we have 
	\vspace{-0.5em}
	\beq
	\lambda_r\sim \text{Gamma}(g_r+s_r,p_r)
	\eeq
	\end{itemize}
	\textbf{Computational Complexity:} Sampling the latent counts $\{\tilde{y}_{\vec{i}r}\}_{r=1}^R$ for each nonzero observation $y_{\vec{i}}$ (note that for $y_{\vec{i}} = 0$, the latent counts are trivially zero) requires computing $\{\zeta_{\vec{i}r}\}_{r=1}^R$, and computing each $\zeta_{\vec{i}r}$ requires $O(K)$ time (Eq~\ref{eq:multsample}). Therefore, sampling all the latent counts $\{\tilde{y}_{\vec{i}r}\}_{r=1}^R$ requires $O(NRK)$ time. Sampling the latent factors $\{\uv_r^{(k)}\}_{r=1}^R$ for the $K$ tensor modes requires $O(RK)$ time. Sampling $\{p_r\}_{r=1}^R$ and $\{\lambda_r\}_{r=1}^R$ requires $O(R)$ time each. Of all these steps, sampling the latent counts $\{\tilde{y}_{\vec{i}r}\}_{r=1}^R$ (which are also used to compute the sufficient statistics $s_{j,r}^{(k)}$ and $s_r$) is the most dominant step, leading to an overall time-complexity of $O(NRK)$ for the Gibbs sampling procedure. 
	
	The linear dependence on $N$ (number of nonzeros) is especially appealing because most real-world count-valued tensors are extremely sparse (have much less than even 1\% nonzeros. In contrast to the standard negative-binomial models for count data, for which the inference complexity also depends on the zeros whose number may be massive (and therefore heuristics, such as subsampling the zeros, are needed), the reparametrizations (Section~\ref{sec:repar}) used by our model allow us to ignore the zeros in the multinomial sampling step (the sufficient statistics do not depend on the zero entries in the tensor), thereby significantly speeding up the inference.  
	\vspace{-0.5em}
	\subsection{Variational Bayes Inference}
	\label{sec:batchvb}
	\vspace{-0.5em}
	Using the mean-field assumption~\cite{jordan1999introduction}, we approximate the target posterior distribution by $Q = \prod_{\iv,r}q(\tilde{y}_{\iv r})\prod_{k,r}q(\uv_r^{(k)})\prod_{r}q(\lambda_r)\prod_{r}q(p_r)$. Our fully conjugate model enables closed-form variational Bayes (VB) inference updates, with the distribution $q(\tilde{y}_{\iv r})$, $q(\uv_r^{(k)})$, $q(\lambda_r)$, and $q(p_r)$ being multinomial, Dirichlet, beta, and gamma, respectively.  We summarize the update equations for the variational parameters of each of these distributions, below:
	\begin{itemize}
	\item \textbf{Updating $\tilde{y}_{\iv r}$:} Using (\ref{eq:multsample}), the updates for $y_{\iv r}$ are given by $\mathbb{E}[y_{\iv r}]=y_{\iv}\zeta_{\vec{i}r}$ where $\zeta_{\vec{i}r}$ is defined as $\zeta_{\vec{i}r}=\frac{\tilde{\zeta}_{\vec{i}r}}{\sum_{r=1}^{R}\tilde{\zeta}_{\vec{i}r}}$
	and $\tilde{\zeta}_{\vec{i}r}$ can be computed as 
	\vspace{-1em}
	\beq
	\tilde{\zeta}_{\vec{i}r}=\exp\{\Psi(s_r+g_r)+\ln(p_r)
	+\sum_{k=1}^{K}\Psi(s_{i_k,r}^{(k)}+a^{(k)})
	-\Psi[\sum_{k=1}^K(s_{i_k,r}^{(k)}+a^{(k)})]\} \label{eq:latcnt}
		\vspace{-0.25em}
	\eeq
	where $\Psi(.)$ is the digamma function, which is the first derivative of the logarithm of the gamma function.\\
	\item \textbf{Updating $\uv_{i_kr}^{(k)}$:} The mean-field posterior $q(\uv_r^{(k)})$ is Dirichlet with each of the component means given by $\mathbb{E}[\uv_{i_kr}^{(k)}]=\frac{\rho_{i_kr}^{(k)}}{\sum_{i_k=1}^{n_k}\rho_{i_kr}^{(k)}}$
	where $\rho_{i_kr}^{(k)}=a^{(k)}+s_{i_k,r}^{(k)}$.\\
	\item \textbf{Updating $p_r$:} The mean-field posterior $q(p_r)$ is beta with mean given by $\mathbb{E}[p_r]= \frac{p_{ra}}{p_{ra}+p_{rb}}$
	where $p_{ra}=c\epsilon+s_r$, $p_{rb}=c(1-\epsilon)+g_r$.\\
	
	\item \textbf{Updating $\lambda_r$:} The mean-field posterior $q(\lambda_r)$ is gamma with mean given by $\mathbb{E}[\lambda_r]=\lambda_{ra}\lambda_{rb}$, where $\lambda_{ra}=(g_r+s_r)$ and $\lambda_{rb}=p_r$.
	\end{itemize}
	
	\textbf{A note on Gibbs vs VB:} The per-iteration time-complexity of the VB inference procedure is also $O(NRK)$. It is to be noted however that, in practice, one iteration of VB in this model is a bit more expensive than one iteration of Gibbs, due to the digamma function evaluation for the $\tilde{\zeta}_{\iv r}$ which is needed in VB when updating the $\tilde{y}_{\iv r}$'s. Prior works on Bayesian inference for topic models~\cite{heinrich2009variational} also support this observation. 

	\vspace{-1em}
	\subsection{Online Inference}
	Batch Gibbs (Section~\ref{sec:batchgibbs}) and VB (Section~\ref{sec:batchvb}) inference algorithms are simple to implement and efficient to run on moderately large-sized problems. These algorithms can however be slow to run for massive data sets (e.g., where the number of tensor entries $N$ and/or the dimension of the tensor is massive). The Gibbs sampler may exhibit slow mixing and the batch VB may be slow to converge. To handle such massive tensor data, we develop two online inference algorithms. The first is online MCMC based conditional density filtering~\cite{guhaniyogi2014bayesian}, while the second is based on stochastic variational inference~\cite{hoffman2013stochastic}. Both these inference algorithms allow processing data in small minibatches and enable our model to analyze massive and/or streaming tensor data. 
	\vspace{-1em}
	\subsubsection{Conditional Density Filtering}
	
	The conditional density filtering (CDF) algorithm~\cite{guhaniyogi2014bayesian} for our model selects a minibatch of tensor entries at each iteration, samples the latent counts $\{\tilde{y}_{\iv r}\}_{r=1}^R$ for these entries conditiond on the previous estimates of the model parameters, updates the sufficient statistics $s_{j,r}^{(k)}$ and $s_r$ using these latent counts (as described below), and resamples the model parameters conditioned on these sufficient statistics. Denoting $I_t$ as data indices in minibatch at round $t$, the algorithm proceeds as
	\begin{itemize}
	\item \textbf{Sampling $\tilde{y}_{\iv r}$:} For all $\iv\in I_t$, sample the latent counts $\tilde{y}_{\iv r(\iv\in I_t)}$ using (\ref{eq:multsample}).\\
	\item \textbf{Updating the conditional sufficient statistics:} Using data from the current minibatch, update the conditional sufficient statistics as: 
	\begin{eqnarray}
	s_{j,r}^{(k,t)} &=& (1-\gamma_t)s_{j,r}^{(k,t-1)} + \gamma_t \frac{N}{B}\sum_{\iv \in I_t:i_k=j}\tilde{y}_{\iv r} \\
	s_r^{(t)} &=& (1-\gamma_t)s_r^{(t-1)} + \gamma_t \frac{N}{B}\sum_{\iv\in I_t}\tilde{y}_{\iv,r}
	\end{eqnarray}
	Note that the updated conditional sufficient statistics (CSS), indexed by superscript $t$, is a weighted average of the old CSS, indexed by superscript $t-1$, and of that computing only using the current minibatch (of size $B$). In addition, the latter term is further weighted by $N/B$ so as to represent the \emph{average} CSS over the \emph{entire} data. In the above,  $\gamma_t$ is defined as $\gamma_t=(t_0+t)^{-\kappa}$, $t_0\ge 0$, and $\kappa\in (0.5,1]$ is needed to guarantee convergence~\cite{cappe2009line}.\\ 
	\item \textbf{Updating $\uv_r^{(k)}, p_r, \lambda_r$:} Using the updated CSS, draw $M$ samples for each of the model parameters $\{\uv_r^{(k,m)}, p_r^{(m)}, \lambda_r^{(m)}\}_{m=1}^M$, from the following conditionals:
	\begin{eqnarray}
	\uv_r^{(k)} &\sim& \text{Dir}(a^{(k)}+s_{1,r}^{(k,t)},\ldots,a^{(k)}+s_{n_k,r}^{(k,t)}) \\
	p_r &\sim& \text{Beta}(c\epsilon+s_r^{(t)},c(1-\epsilon)+g_r) \\
	\lambda_r &\sim& \text{Gamma}(g_r+s_r^{(t)},p_r)
	\end{eqnarray}
	and either store the sample averages of $\uv_r^{(k)}, p_r$, and $\lambda_r$, or their analytic means to use for the next CDF iteration~\cite{guhaniyogi2014bayesian}. Since the analytic means of the model parameters are available in closed-form in this case, we use the latter option, which obviates the need to draw $M$ samples, thereby also speeding up the inference significantly.
	\end{itemize}
	We next describe the stochastic (online) VB inference for our model.
	\vspace{-1em}
	\subsubsection{Stochastic Variational Inference}
	The batch VB inference (Section~\ref{sec:batchvb}) requires using the entire data for the parameter updates in each iteration, which can be computationally expensive and can also result in slow convergence. 
	Stochastic variational inference (SVI), on the other hand, leverages ideas from stochastic optimization~\cite{hoffman2013stochastic} and, in each iteration, uses a small randomly chosen minibatch of the data to updates the parameters. Data from the current minibatch is used to compute stochastic gradients of the variational objective w.r.t. each of the parameters and these gradients are subsequently used in the parameter updates. For our model, the stochastic gradients depend on the sufficient statistics computed using the current minibatch $I_t$: $s_{j,r}^{(k,t)}=\sum_{\iv \in I_t:i_k=j}\tilde{y}_{\iv r}$ and $s_r^{(t)}=\sum_{\iv\in I_t}\tilde{y}_{\iv,r}$, where $\tilde{y}_{\iv r}$ is computed using Eq~\ref{eq:latcnt}. Denoting $B$ as the minibatch size, we reweight these statistics by $N/B$ to compute the \emph{average} sufficient statistics over the entire data~\cite{hoffman2013stochastic} and update the other variational parameters as follows:
\begin{eqnarray}
	\rho_{i_kr}^{(k,t)} &=& (1-\gamma_t)\rho_{i_kr}^{(k,t-1)} + \gamma_t (a^{(k)}+(N/B)s_{i_k,r}^{(k,t)}) \\
p_{ra}^{(t)} &=& (1-\gamma_t)p_{ra}^{(t-1)} + \gamma_t (c\epsilon + (N/B)s_r^{(t)})	\\
p_{rb}^{(t)} &=& (1-\gamma_t)p_{rb}^{(t-1)} + \gamma_t (c(1-\epsilon) + g_r)	\\
\lambda_{ra}^{(t)} &=& (1-\gamma_t)\lambda_{ra}^{(t-1)} + \gamma_t (g_r+(N/B)s_r^{(t)})\\
\lambda_{rb}^{(t)} &=& (1-\gamma_t)\lambda_{ra}^{(t-1)} + \gamma_t p_r
\end{eqnarray}	
where $\gamma_t$ is defined as $\gamma_t=(t_0+t)^{-\kappa}$, $t_0\ge 0$, and $\kappa\in (0.5,1]$ is needed to guarantee convergence~\cite{hoffman2013stochastic}. 	
\vspace{-0.5em}
\subsubsection{Computational Complexity:}
In contrast to the batch Gibbs and batch VB, both of which have $O(NRK)$ cost per-iteration, the per-iteration cost of the online inference algorithms (CDF and SVI) is $O(|I_t|RK)$ where $|I_t|$ is the minibatch size at round $t$. We use a fixed minibatch size $B$ for each minibatch, so the per-iteration cost is $O(BRK)$.

	\vspace{-1em}
	\section{Related Work}
	\label{sec:relwork}
	\vspace{-0.5em}
	Although tensor factorization methods have received considerable attention recently, relatively little work exists on scalable analysis of massive count-valued tensor data.
Most of the recently proposed methods for scalable tensor decomposition~\cite{kang2012gigatensor,beutel2014flexifact,inah2015haten2,papalexakis2015parcube} are based on minimizing the Frobenious norm of the tensor reconstruction	error, which may not be suitable for count or overdispersed count data. The rank of decomposition also needs to be pre-specified, or chosen via cross-validation. Moreover, these methods assume the tensor to be fully observed and thus cannot be used for tensor completion tasks. Another key difference between these methods and ours is that scaling up these methods requires parallel or distributed computing infrastructure, whereas our fully Bayesian method exhibits excellent scalability on a single machine. At the same time, the simplicity of the inference update equations would allow our model to be easily parallelized or distributed. We leave this possibility to future work. 
	
One of the first attempts to explicitly handle count data in the context of non-negative tensor factorization includes the work of~\cite{chi2012tensors}, which is now part of the Tensor Toolbox~\footnote{\url{http://www.sandia.gov/~tgkolda/TensorToolbox/index-2.6.html}}. This method optimizes the Poisson likelihood, using an alternating Poisson regression sub-routine, with non-negative constraints on the factor matrices. However, this method requires the rank of the decomposition to be specified, and cannot handle missing data. Due to its inability in handling missing data, for our experiments (Section~\ref{sec:expt}), as a baseline, we implement and use a Bayesian version of this model which \emph{can} handle missing data.

Among other works of tensor factorization for count data, the method in~\cite{tensoricassp} can deal with missing values, though the rank still needs to be specified, and moreover the factor matrices are assumed to be real-valued, which makes it unsuitable for interpretability of the inferred factor matrices.

In addition to the Poisson non-negative tensor factorization method of~\cite{chi2012tensors}, some other non-negative tensor factorization methods~\cite{shashua2005non,cichocki2009nonnegative,schmidt2009probabilistic} also provide interpretability for the factor matrices. However, these methods usually have one or more of the following limitations: (1) there is no explicit generative model for the count data, (2) the rank needs to be specified, and (3) the methods do not scale to the massive tensor data sets of scales considered in this work.

Methods that facilitate a full Bayesian analysis for massive count-valued tensors, which are becoming increasingly prevalent nowadays, are even fewer. A recent attempt on Bayesian analysis of count data using Poisson likelihood is considered in~\cite{schein2014}; however, unlike our model, their method cannot infer the rank and relies on batch VB inference, limiting its scaling behavior. Moreover, the Poisson likelihood may not be suitable for overdispersed counts.	
	
	Finally, inferring the rank of the tensor, which is NP-complete in general~\cite{kolda2009tensor}, is another problem for which relatively little work exists. Recent attempts at inferring the rank of the tensor in the context of CP decomposition include~\cite{rai14tensor,zhao2014bayesian}; however (1) these methods are not applicable for count data, and (2) the inferred factor matrices are real-valued, lacking the type of interpretability needed in many applications.
	
Our framework is similar in spirit to the matrix factorization setting proposed in~\cite{zhou2012NBPFA} which turns out to be a special case of our framework. In addition, while~\cite{zhou2012NBPFA} only developed (batch) Gibbs sampling based inference, we present both Gibbs sampling as well as variational Bayesian inference, and design efficient \emph{online} Bayesian inference methods to scale up our framework for handling massive real-world tensor data.
	
To summarize, in contrast to the existing methods for analyzing tensors, our fully Bayesian framework, based on a proper generative model, provides a flexible method for analyzing massive count-valued tensors, side-stepping crucial issues such as rank-specification, providing good interpretability of the latent factors, while still being scalable for analyzing massive real-world tensors via online Bayesian inference.
		
	\vspace{-1em}	
	\section{Experiments}
	\label{sec:expt}
	\vspace{-1em}	
	We apply the proposed model on a synthetic and three real-world data sets that range in their sizes from moderate to medium to massive. The real-world tensor data sets we use in our experiments are from diverse application domains, such as analyzing country-country interaction data in political science, topic modeling on \emph{multiway} publications data (with entities being authors, words, and publication venues), and analysis of massive household transactions data. These data sets include:
	
	\begin{itemize}
		\item \textbf{Synthetic Data:} This is a tensor of size $300\times 300 \times 300$ generated using our model by setting an upper bound $R = 50$ over the number of factors; only 20 factors were significant (based on the values of $\lambda_r$), resulting in an effective rank 20.\\
		\item \textbf{Political Science Data (GDELT):} This is a real-world four-way tensor data of country-country interactions. The data consists of 220 countries, 20 action types, and the interactions date back to 1979~\cite{leetaru2013gdelt}. We focus on a subset of this data collected during the year 2011, resulting in a tensor of size $220\times 220\times 20\times 52$. Section~\ref{sec:polsc} provides further details.\\ 
		\item \textbf{Publications Data:} This is a $2425 \times 9088 \times 4068$ count-valued tensor,  constructed from a database of research papers published by researchers at Duke University\footnote{Obtained from \url{https://scholars.duke.edu/}}; the three tensor modes correspond to authors, words, and venues. Section~\ref{sec:scholardata} provides further details.\\
		\item \textbf{Transactions (Food) Data:} This is a $117054\times 438 \times 67095$  count-valued tensor, constructed from a database of transactions data of food item purchases at various stores in the US~\footnote{Data provided by United States Department of Agriculture (USDA) under a Third Party Agreement with Information Resources, Inc. (IRI)}; the three tensor modes correspond to households, stores, and items. Section~\ref{sec:transactionsdata} provides further details.
	\end{itemize}
	
	
	We compared our model with the following baselines: ($i$) Bayesian Poisson Tensor Factorization (\textsc{BayesPTF}), which is fully Bayesian version of the Poisson Tensor Factorization model proposed in~\cite{chi2012tensors}, and ($ii$) Non-negative Tensor Decomposition based on Low-rank Approximation (\textsc{lraNTD}) proposed in~\cite{lraNTD}. All experiments are done on a standard desktop computer with Intel i7 3.4GHz processor and 24GB RAM.
	
	\vspace{-1em}	
	\subsection{Inferring the Rank}
	\label{sec:synth}
\begin{figure}[!htbp]
			\vskip -0.5in
			\begin{center}
				\centerline{\includegraphics[trim = 48mm 10mm 23cm 45mm,clip,scale=0.27]{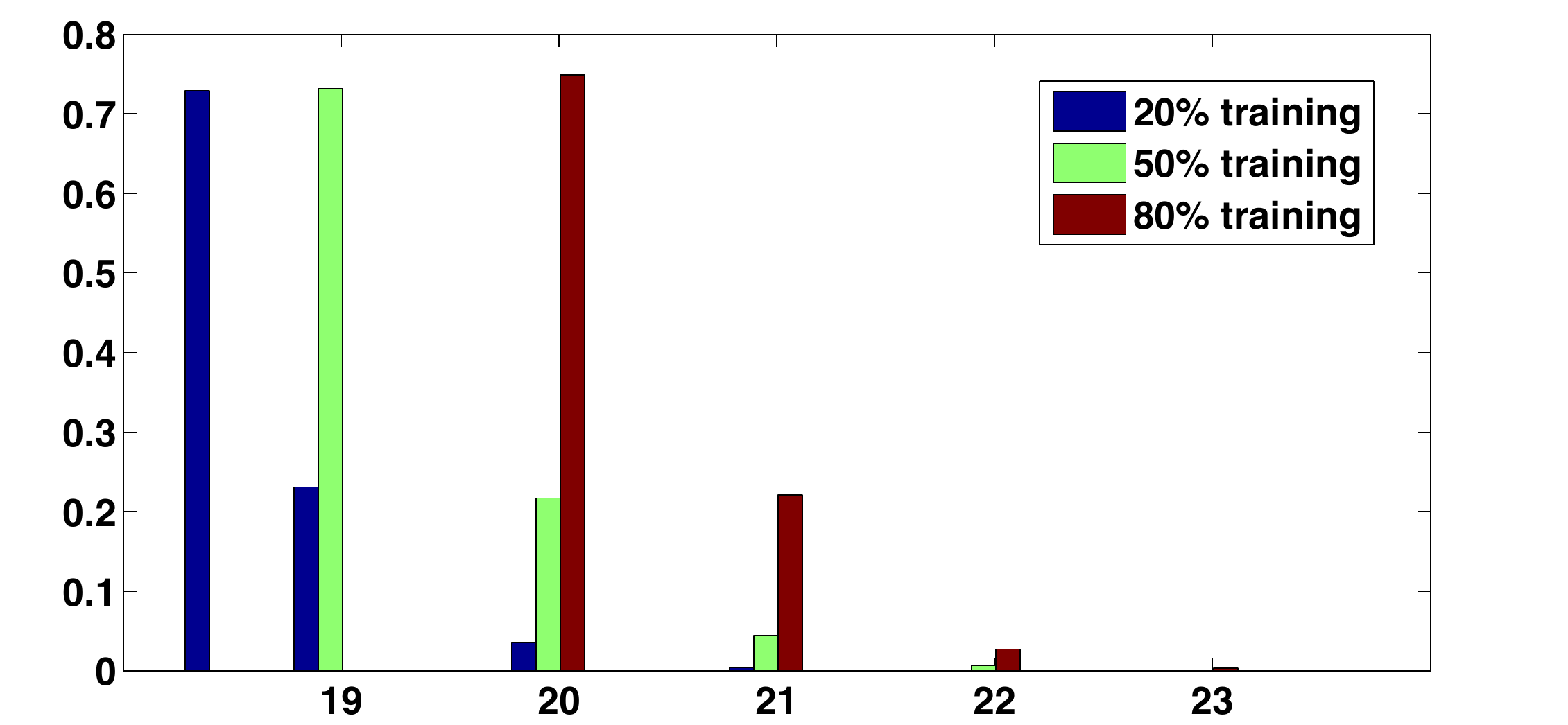}\includegraphics[trim = 30mm 90mm 0cm 55mm,clip,scale=0.28]{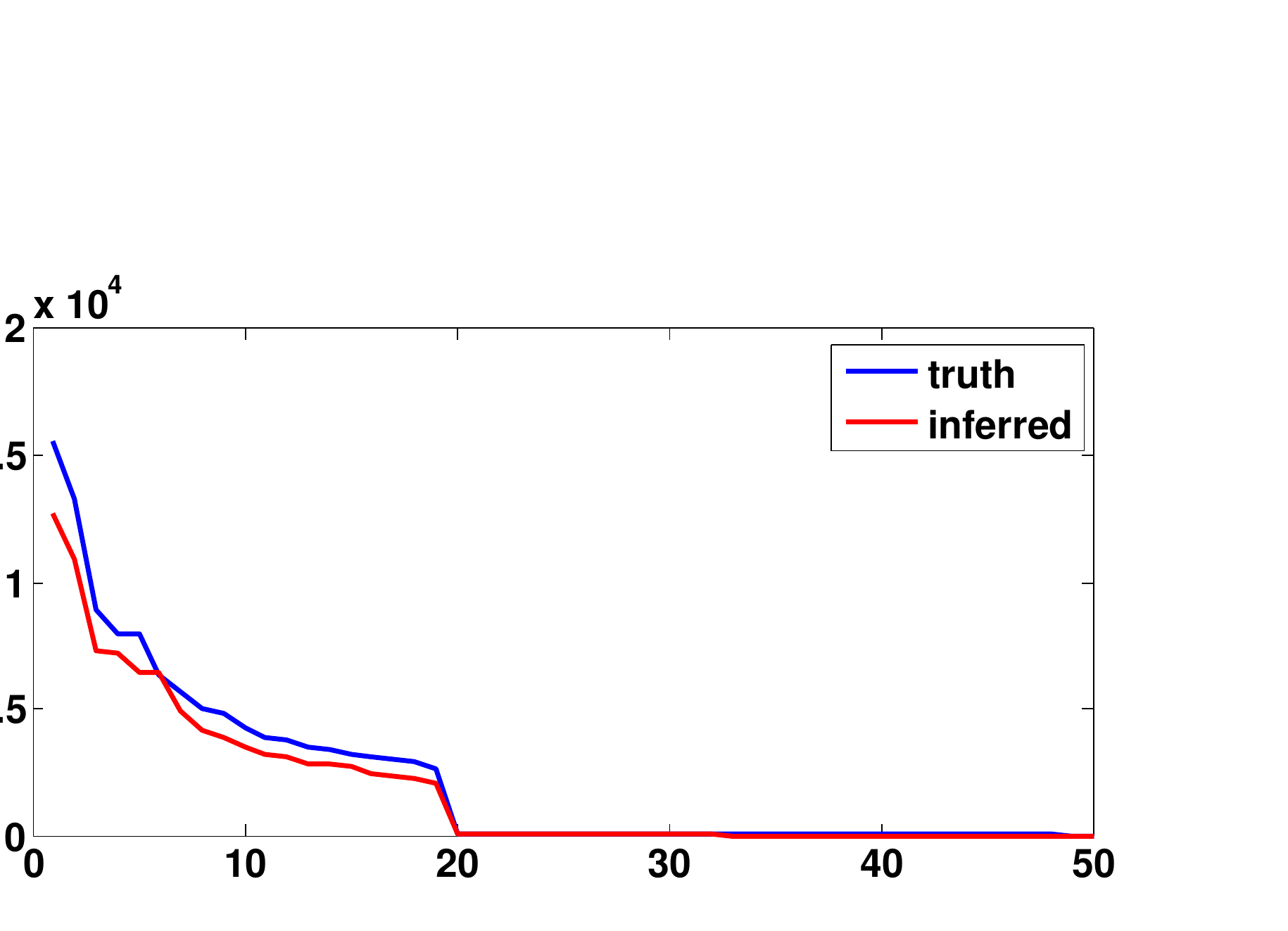}}
				\vspace{-1em}
				\caption{Distribution over inferred ranks for syntheric data (left), and $\lambdav$ inferred using 80\% training data (right).}
				\label{fig:rankhist}
			\end{center}
			\vskip -0.5in
	\end{figure}

 	To begin with, as a sanity check for our model, we first perform an experiment on the synthetic data described above to see how well the model can recover the true rank (tensor completion results are presented separately in Section~\ref{sec:completion}). 
 	For this experiment, we run the batch Gibbs sampler (the other inference methods also yield similar results) with 1000 burn-ins, and 1000 collection samples. We experiment with three settings: using 20\%, 50\% and 80\% data for training. The empirical distribution (estimated using the collected MCMC samples) of the effective inferred rank for each of these settings is shown in Figure~\ref{fig:rankhist} (left). In each collection iteration, the effective rank is computed after a simple thresholding on the $\lambda_r$'s where components with very small $\lambda_r$ are not counted (also see Figure~\ref{fig:rankhist} (right)). With 80\% training data, the distribution shows a distinct peak at 20 and even with smaller amounts of training data (20\% and 50\%), the inferred rank is fairly close to the ground truth of 20. In Figure~\ref{fig:rankhist} (right), we show the spectrum of all the $\lambda_r$'s comparing the ground truth vs the inferred values; 
 	
		\vspace{-1em}
\subsection{Tensor Completion Results}
\label{sec:completion}

We next experiment on the task of tensor completion, where for each method 95\% of the data are used for training and the remaining 5\% data is used as the heldout set (note that the data sets we use are extremely sparse in nature, with considerably less than 1\% entries of the tensor being actually observed). The results are reported in Table~\ref{tab:comp} where we show the log likelihood and the mean-absolute error (MAE) in predicting the heldout data. Timing-comparison for the various batch and online inference methods is presented separately in Section~\ref{sec:scale}. 

For this experiment, we compare our \textsc{BNBCP} model (using the various inference methods) with (1) \textsc{BayesPTF} - a fully Bayesian variant (we implented it ourselves) of a state-of-the-art Poisson Tensor Factorization model originally proposed in~\cite{chi2012tensors} (which cannot however handle missing data), and (2) \textsc{lraNTD}~\cite{lraNTD} which is an optimization based non-negative tensor decomposition method. As Table~\ref{tab:comp} shows, our methods achieve better log-likelihood and MAE as compared to these baselines. Moreover, among our batch and online Bayesian inference methods, the online inference methods give competitive or better results as compared to their batch counterparts. In particular, the online MCMC method based on conditional density filtering (\textsc{BNBCP-CDF}) works the best across all the methods (please see Section ~\ref{sec:scale} for a timing comparison). 

 		\begin{table*}[!htbp]
			
			\vskip -0.15in
			\begin{center}
				\begin{scriptsize}
					\begin{sc}
						\begin{tabular}{l|cccc|cccc}
							\hline
							Datasets & Toy data & GDELT &  Publication & Food& Toy data & GDELT &  Publication & Food\\
							\hline
					BayesPTF       & -107563 &      -4425695     &-860808 &   -2425433       & 1.012 &    55.478    & 1.636 & 1.468\\
					lraNTD        &    N/A       &      N/A     &    N/A   &  N/A  & 1.019   &    65.049    &    N/A   & N/A \\					
					\hline
					BNBCP-Gibbs    & -97580  &  -3079883         &-619258 & -2512112 & 0.989 &  45.436      & 1.565 & 1.459 \\
					BNBCP-VB       & -99381  &     -2971769      &-632224 & -2533086 & 0.993 &   \textbf{43.485}     & 1.574 & 1.472\\
					\hline
					BNBCP-CDF     & -95472  &    \textbf{-2947309}       &\textbf{-597817} & \textbf{-2403094}         & 0.985 &   44.243     & \textbf{1.555} & \textbf{1.423} \\
					BNBCP-OnlineVB & -98446  &      -3169335     &-660068 &  -2518996        & 0.989 &    46.188    & 1.601 & 1.461\\
							\hline
						\end{tabular}
					\end{sc}
					\vspace{2em}
			\caption{Loglikelihood and MAE comparison for different methods (the two baselines, our model with batch inference, and our model with online inference) on four datasets. Note: \textsc{lraNTD} gave out-of-memory error on publications and food transactions data sets so we are unable to report its results on these data sets. We also only report the MAE for \textsc{lraNTD}, and not the log-likelihood,  because it uses a Gaussian likelihood model for the data.} 	
							\label{tab:comp}
				\end{scriptsize}
			\end{center}
			\vskip -0.5in
		\end{table*}
	\vspace{-1em}
	\subsection{Analyzing Publications Database}
	\label{sec:scholardata}
	The next experiment is on a three-way tensor constructed from a scientific publications database. The data consist of abstracts from papers published by various researchers at Duke University~\footnote{Data crawled from \url{https://scholars.duke.edu/}}. In addition to the paper abstract, the venue information for each paper is also available. The data collection contains 2425 authors, 9088 words (after removing stop-words), and 4068 venues which results in a $2425 \times 9088 \times 4068$ word-counts tensor, on which we run our model. As the output of the tensor decomposition, we get three factor matrices. Since the latent factors in our model are non-negative and sum to one, each latent factor can also be interpreted as a \emph{distribution} over authors/words/venues, and consequently represents a ``topic''. Therefore the three factor matrices inferred by our model for this data correspond to authors $\times$ topics, words $\times$ topics, and venue $\times$ topics, which we use to further analyze the data.
	
	We apply the model \textsc{BNBCP-CDF} on this data (with $R=200$) and using the inferred words $\times$ topics matrix, in Table \ref{tab:topic} (left) we show the list of 10 most probable words in four factors/topics that seem to correspond to \textbf{optics}, \textbf{genomics}, \textbf{machine learning \& signal processing}, and \textbf{statistics}. To show the topic representation across different departments, we present a histogram of \emph{departmental affiliations} for 20 authors with highest probabilities in these four factors. We find that, for the genomics factor, the top authors (based on their topic scores) have affiliations related to biology which makes intuitive sense. Likewise, for the statistics factor, most of the top authors are from statistics and biostatistics departments. The top 20 authors in factors that correspond to optics and machine learning \& signal processing, on the other hand, are from departments of electrical and computer engineering and/or computer science, etc.
	
		\begin{table*}[!htbp]
			\vskip -0.15in
			\begin{center}
				\begin{scriptsize}
					\begin{sc}
						\begin{tabular}{llll|l}
							\hline
							Optics & Genomics & ML/SP & Stats & Top Venues in ML/SP\\
							\hline
							gigapixel     & gene         & dictionary & model         & ICASSP\\
							microcamera   & chromatin    & sparsity   & priors        & IEEE trans. sig. proc.\\
							cameras       & occupancy    & model      & bayesian      & ICML\\
							aperture      & centromere   & bayesian   & lasso         & Siam J. img. sci.\\
							lens          & transcription& compressed & latent        & IEEE trans. img. proc.\\
							multiscale    & genome       & compressive& inference     & IEEE int. symp. biomed. img.\\
							optical       & sites        & matrix     & regression    & NIPS\\
							system        & expression   & denoising  & sampler       & IEEE trans. wireless comm.\\
							nanoprobes    & sequence     & gibbs      & semiparametric& IEEE workshop stat. sig. proc.\\
							metamaterial  & vegfa        & noise      & nonparametric & IEEE trans. inf. theory\\
							\hline
						\end{tabular}
					\end{sc}
					\vspace{1em}
			\caption{Most probable words in topics related to optics, genomics, machine learning/signal processing(ML/SP) and statistics (Stats), and top ranked venues in ML/SP community.}					\label{tab:topic}
				\end{scriptsize}
			\end{center}
			\vskip -0.5in
		\end{table*}		
	
		\begin{figure}[!htbp]
		\vskip -0.35in
		\begin{center}
			\centerline{
				\includegraphics[scale=0.27,angle=90]{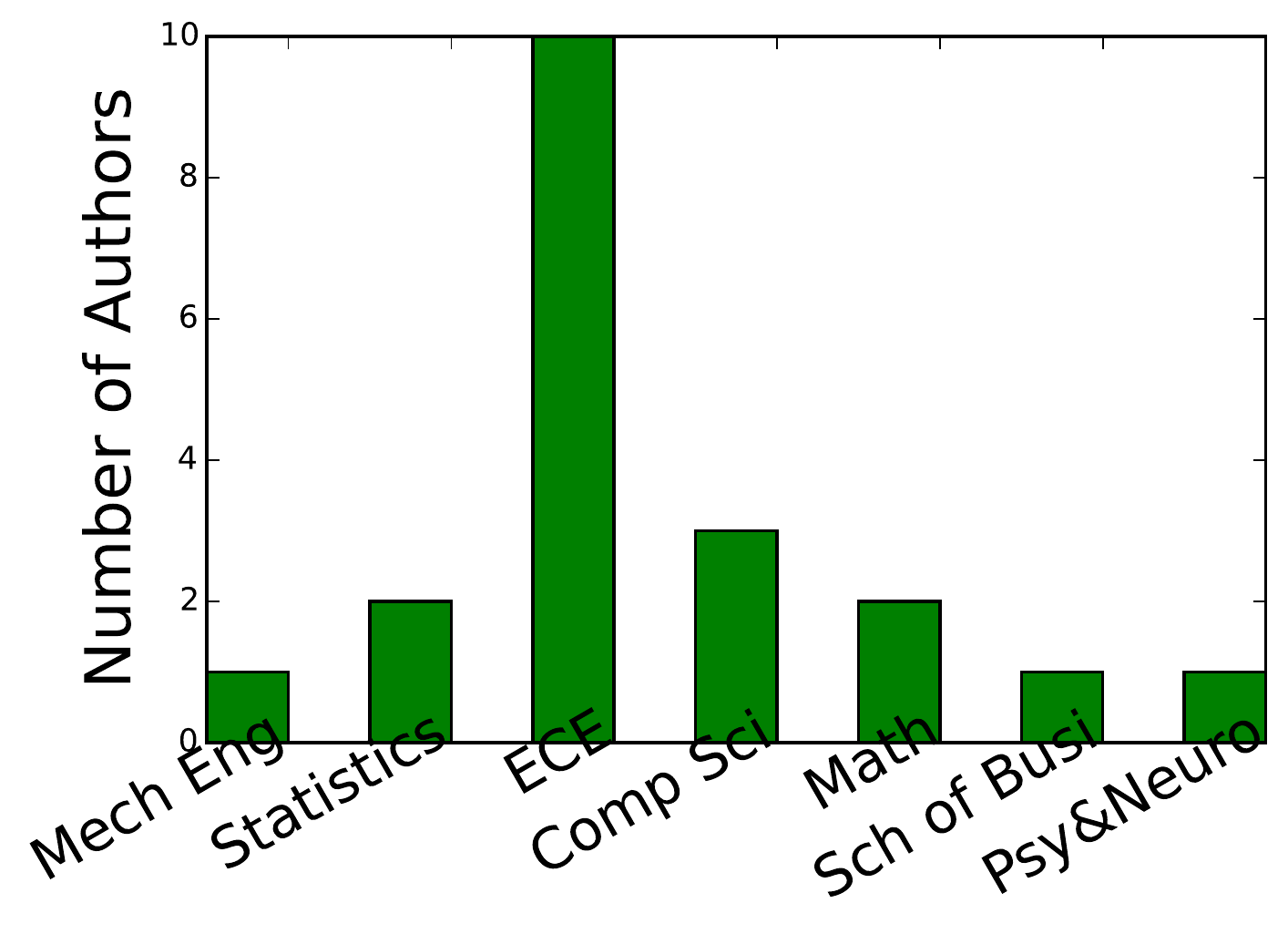}
				\includegraphics[scale=0.27,angle=90]{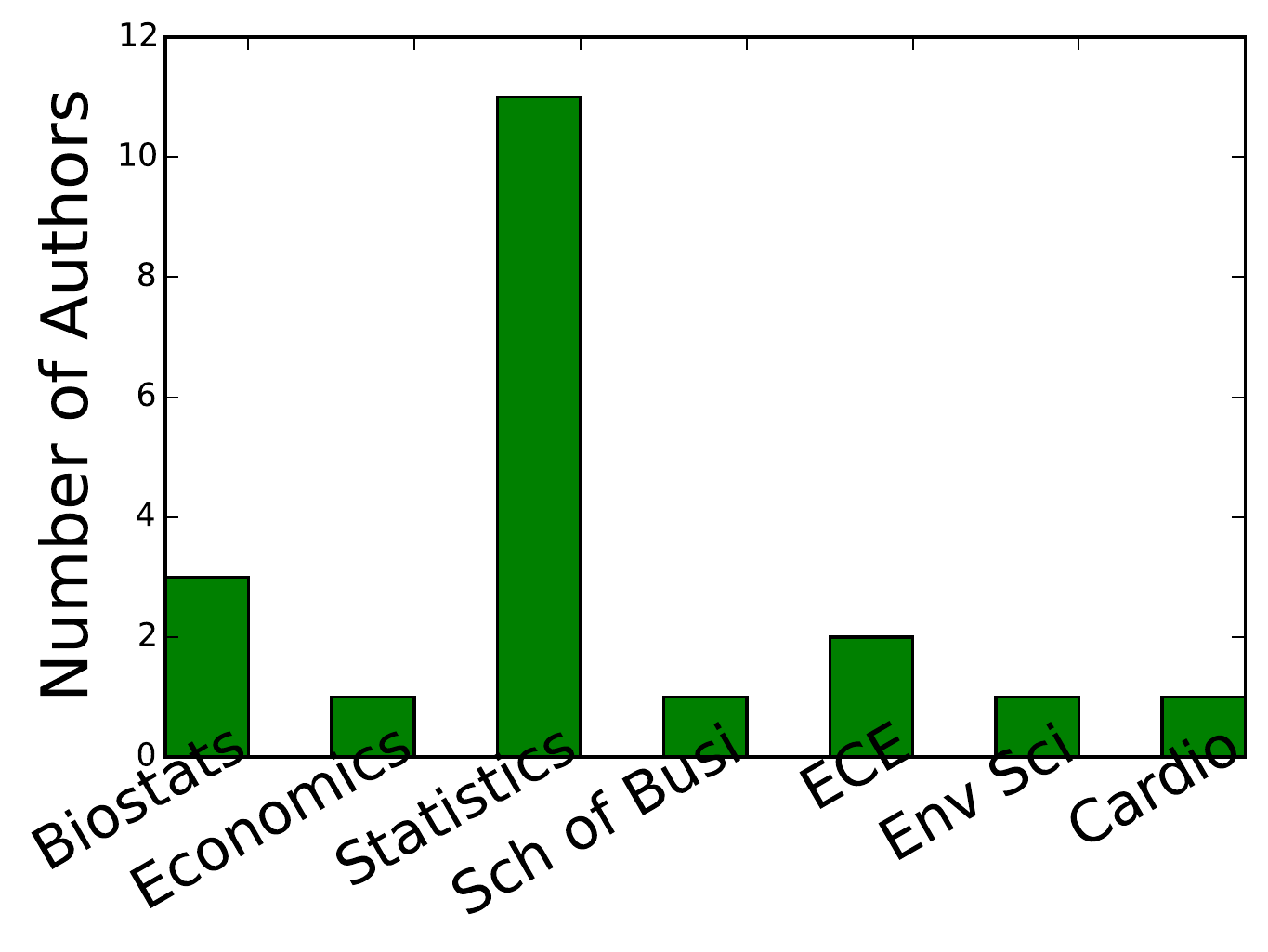}
			}
			\centerline{
				\includegraphics[scale=0.27,angle=90]{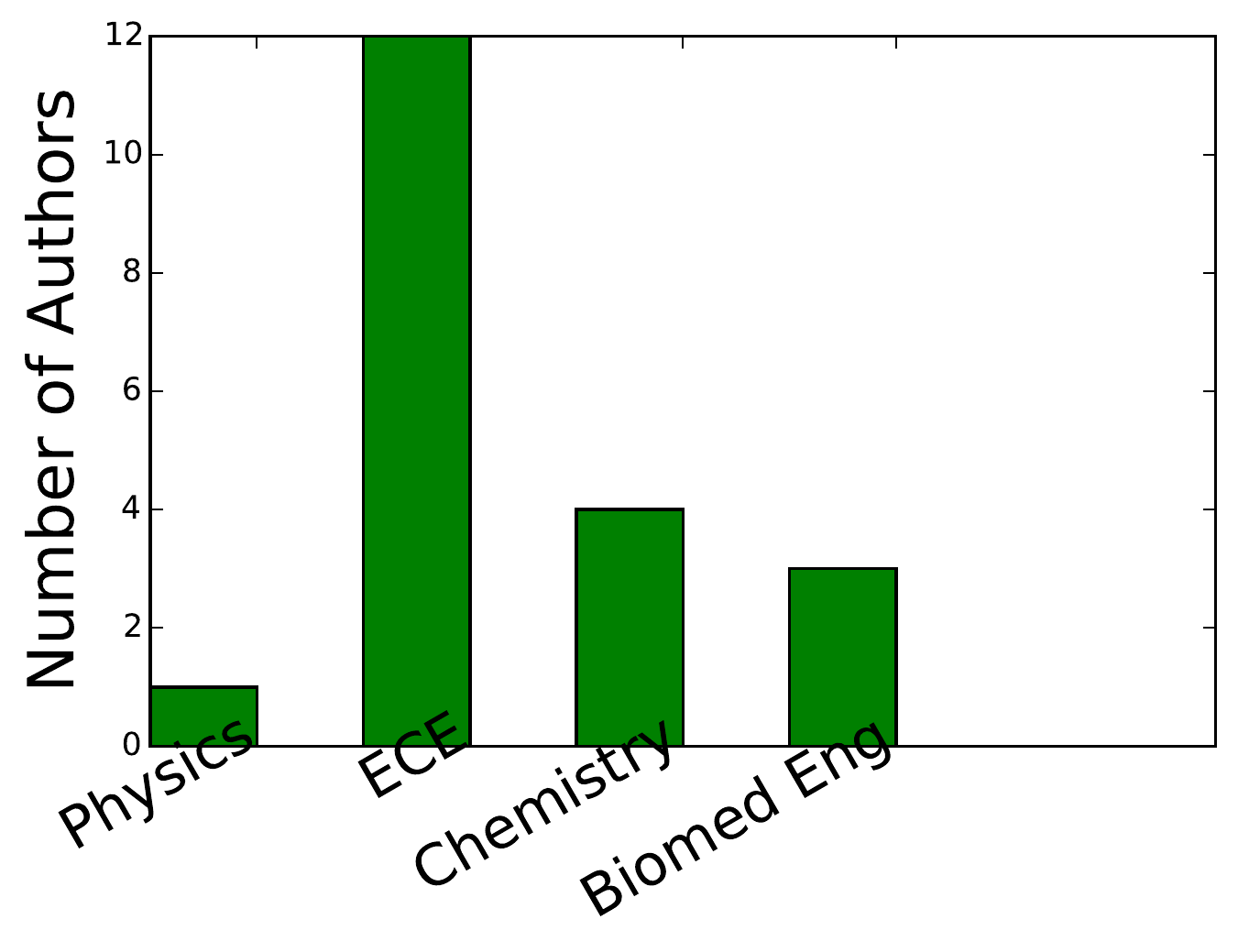}
				\includegraphics[scale=0.27,angle=90]{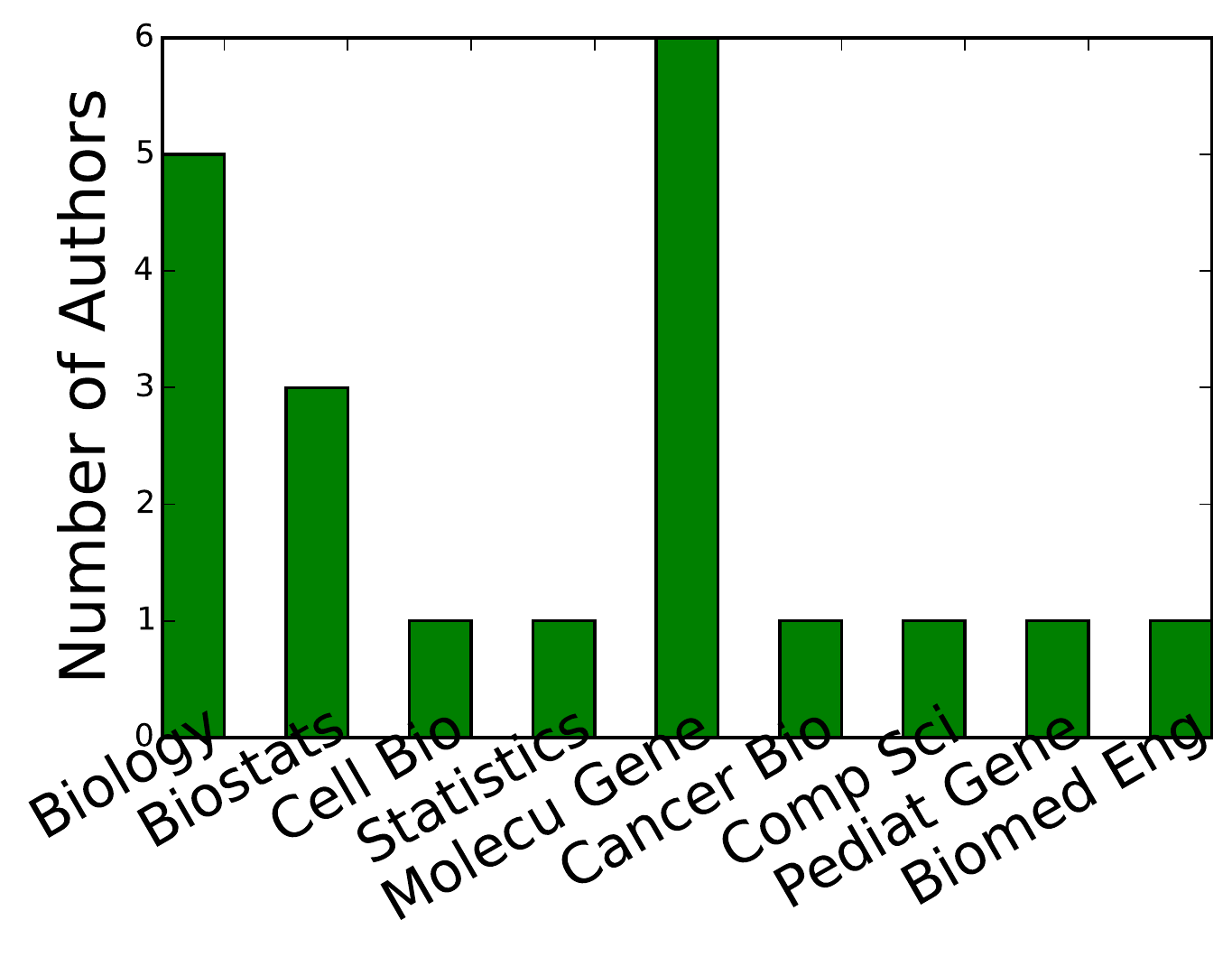}
				}
			\vskip -1em
			\caption{Histogram of affiliations for top 20 authors in factors related to machine learning/signal processing (top left) and statistics (top right), optics (bottom left), and genomics(bottom right)}
			\label{fig:hist}
		\end{center}
		\vskip -0.5in
	\end{figure}	
	
	Similarly, using the inferred venues $\times$ topics matrix, we list the most likely venues for each topic. Due to space-limitations, here we only present the most likely venues in machine learning \& signal processing factor/topic; the result is shown in Table \ref{tab:topic} (right-most column). The result shows that venues like ICASSP, IEEE Trans. Signal Proc., ICML, and NIPS all rank at the top in the machine learning \& signal processing factor, which again makes intuitive sense.
	
				%
					
	\subsection{Analyzing Political Science Data}
	\label{sec:polsc}
	We use the model to analyze a real-world political science data set consisting of country-country interactions. Such analyses are typically done by political scientists to study, analyze and understand complex international multilateral relations among countries.
	The data set is from the Global Database of Events, Location, and Tone (GDELT) ~\cite{leetaru2013gdelt}. GDELT records the dyadic interactions between countries in the form of ``Country A did something to Country B''. In our experiments, we consider 220 countries (``actors'') and 20 unique high-level action types in 52 weeks of year 2012. After preprocessing, we have a four-way (country-country-action-time) action counts tensor of size $220\times 220 \times 20 \times 52$. Note that both first and second tensor mode represents countries; first mode as ``sender'' and the second mode as ``receiver'' of a particular action. In this analysis, we set $R$ to be large enough (200) and the model discovered roughly about 120 active components (i.e., components with significant value of $\lambda_r$).
	
	\begin{figure}[!htbp]
	\vskip -0.2in
	\begin{center}
		\centerline{\includegraphics[trim = 35mm 0mm 23cm 0mm,clip,scale=0.32]{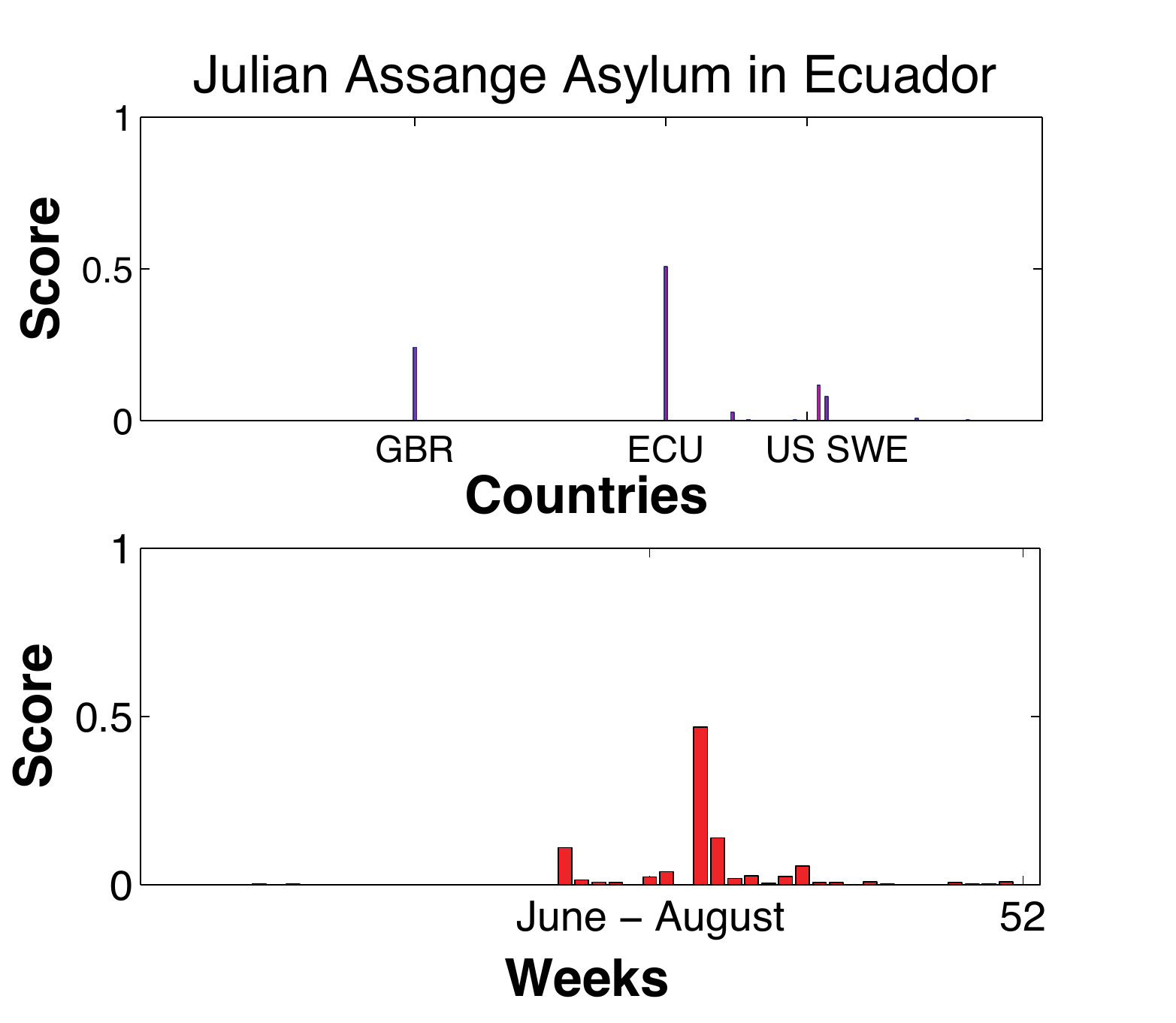}\includegraphics[trim = 35mm 0mm 23cm 0mm,clip,scale=0.32]{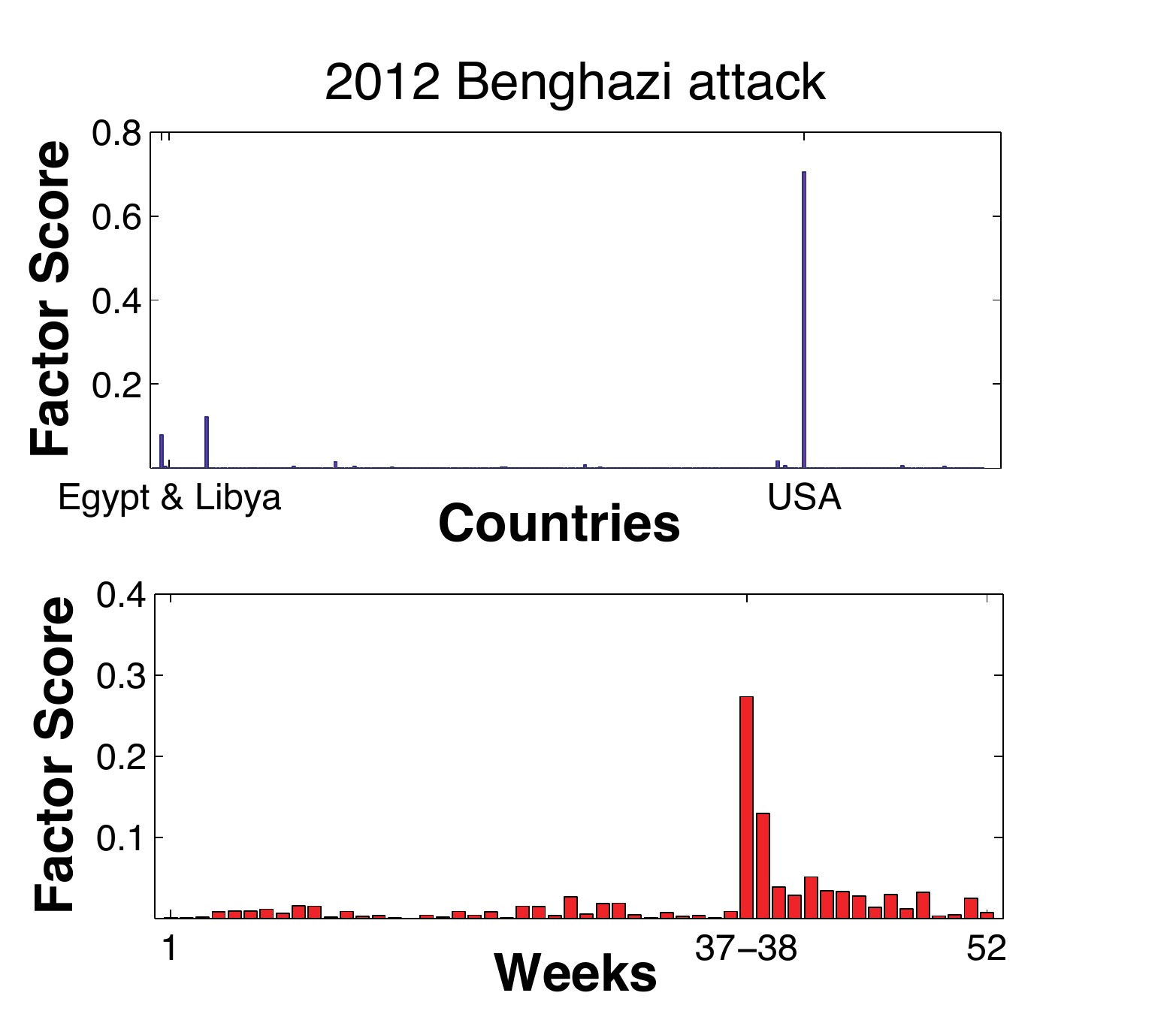}}
		\caption{Country factors (top row) and time factors (bottom row) for Julian Assange asylum in Ecuador (left column) and 2012 Benghazi attack (right column).}
		\label{fig:gdelt}
	\end{center}
	\vskip -0.5in
	\end{figure}
	
	We apply the model (\textsc{BNBCP-CDF}; other methods yield similar results) and examine each of the time dimension factors, specifically looking for the significant components (based on the magnitude of $\lambda_r$) in which the time dimension factor also peaks during certain time(s) of the year. We show results with two such factors in Figure~\ref{fig:gdelt}. In Figure~\ref{fig:gdelt} (column 1), the time and country (actor) factors seems to suggest that this factor/topic corresponds to the event ``Julian Assange''. The actor subplot shows spikes at Ecuador, United Kingdom, United States, and Sweden whereas the time factor in the bottom left subplot shows spikes between June and August. The time and countries involved are consistent with the public knowledge of the event of Julian Assange seeking refuge in Ecuador.
	
	Likewise, in Figure~\ref{fig:gdelt} (column 2), the time and country (actor) factors seems to suggest that this factor corresponds to the event ``Benghazi Attack'' which took place on Sept. 12 (week 37) of 2012, in which Islamic militants attacked American diplomatic compound in Benghazi, Libya. The attack killed an US Ambassador. As the Figure shows, the top actors identified are US, Libya and Egypt, and spikes are found at around week 37 and 38, which are consistent with the public knowledge of this event. 
	
	The results of these analyses demonstrate that the interpretability of our model can be useful for identifying events or topics in such multiway interaction data.
	
	\subsection{Analyzing Transactions Data}
	\label{sec:transactionsdata}
        We next apply our model (\textsc{BNBCP-CDF}; other methods yield similar results) for analyzing transactions data for food item purchases made at stores. Our data is collected for a demographically representative sample of US consumers who reside in large urban and suburban areas and purchase food in supermarkets and grocery stores. The data were provided by the USDA under a Third Party Agreement with IRI. Each transaction is identified by a unique Universal Product Code (UPC) barcode and the store where the transaction occurred. Some items such as fresh produce do not have UPCs and are identified separately. The households are observed over a four year period, during which they are provided with a technology that allows them to scan each purchase and record additional information such as the store where the purchase was made (and other economic data). Participating households are provided with incentives designed to encourage compliance. For each household-product-store combination we record the number of unique purchases over the sampling period. The database has a total of 117,054 unique households, 438 stores, and 67,095 unique items and we construct a 3-way count tensor of size $117054\times 438 \times 67095$ with about 6.2 million nonzero entries.
        
        We apply the proposed model on this data by setting $R=100$ (out of which about 60 components were inferred to have a significant value of $\lambda_r$) and looked at the stores factor matrix. Since each column (which sums to 1) of the store factor matrix can be thought of as a \emph{distribution} over the stores, we look at three of the factors from the store factor matrix and tried to identify the stores that rank at the top in that factor. In Table~\ref{tab:store}, we show results from each of these factors. Factor 1 seems to suggest that it is about the most popular stores (included Walmart, for example), Factor 2 has stores that primarily deal in wholesale (e.g., Costco, Sam's Wholesale Club), and Factor 3 contains stores that sell none or very few food items (e.g., Mobil, Petco). Note that the Walmart Super Center figures prominently in both Factor 1 and Factor 2.
        
	\begin{table}[!htbp]
		\vskip -0.15in
		\begin{center}
			\begin{scriptsize}
				\begin{sc}
					\begin{tabular}{lll}
						\hline
						Factor 1 & Factor 2 & Factor 3 \\
						\hline
						Walmart Sup. Center & Sam's Club & Dick's Sporting\\
						Walmart Traders & Meijer & Mobil\\
						Walmart Neighb. & Costco & Petco\\
						Walmart & B J'S Wholesale & Sally Beauty\\
						Kroger & Walmart Sup. Center & GNC All\\
						\hline
					\end{tabular}
				\end{sc}
			\end{scriptsize}
			\vspace{1em}
		\caption{Three of the store factors inferred from the transaction data (top-5 stores shown for each)}	
				\label{tab:store}
		\end{center}
		\vspace{-2em}
	\end{table}
	\vspace{-2em}
	\begin{figure}[!htbp]		
		\begin{center}
			\centerline{\includegraphics[scale=0.2]{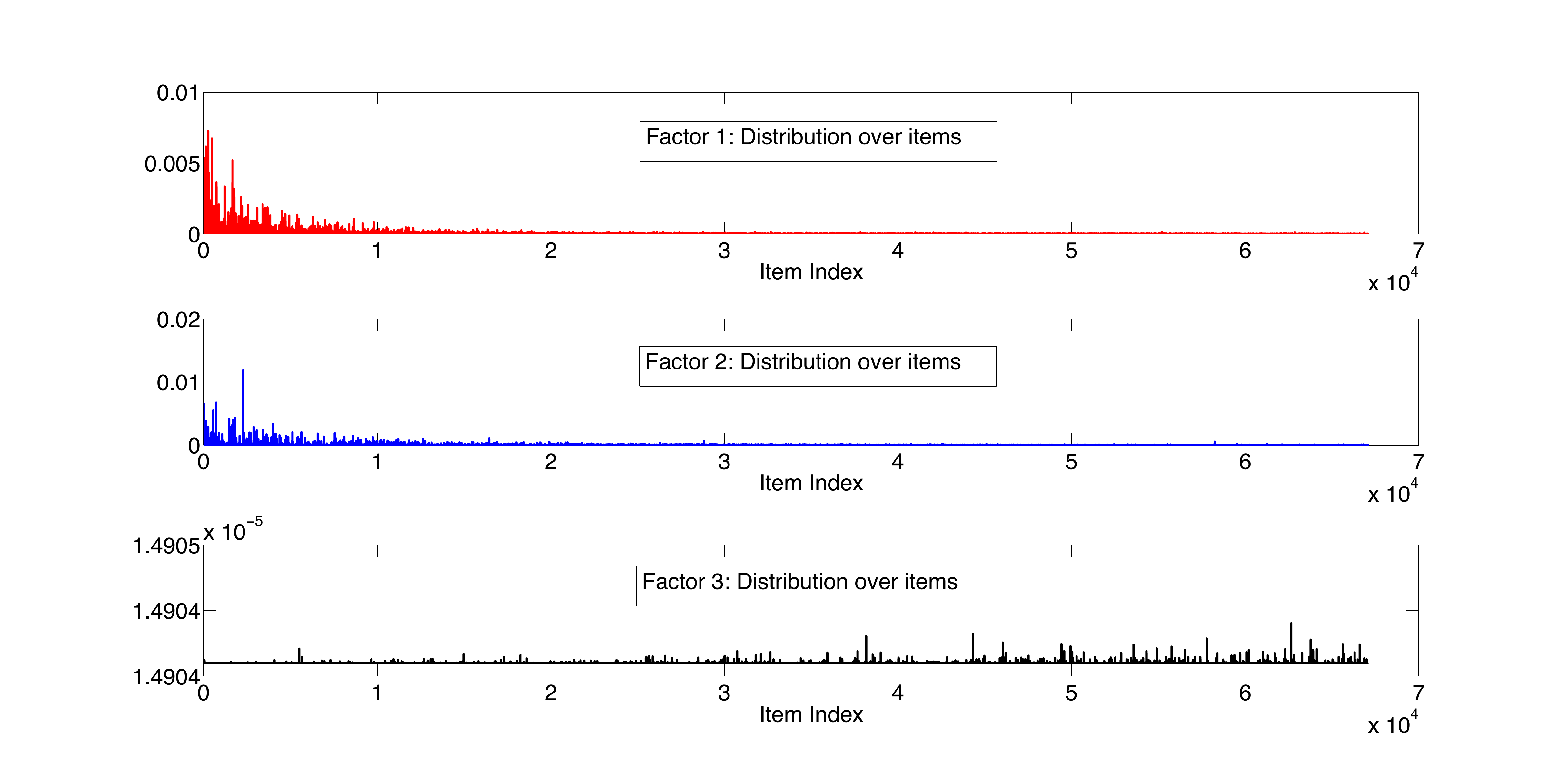}}
			\vskip -0.2in
			\caption{Distributions over items for three factors (each factor corresponds to a cluster).}
			\label{fig:itemclusters}
		\end{center}
		\vskip -0.5in		
	\end{figure}
        We next look at the items factor matrix. In Figure~\ref{fig:hist}, we plot the inferred distribution over items in each of the three clusters described above. For factors 1 and 2 (which correspond to the most popular stores and wholesale stores respectively), the distribution over the items (top and bottom panel in Figure~\ref{fig:hist}) have a reasonably significant mass over a certain range of items (for the items indexed towards the left side in the plots of factors 1 and 2). On the other hand, for factor 3 which corresponds to stores that sell no or very few types of food items, the distribution over the items is rather flat and diffuse with very weak intensities (looking at the scale on the y axis). From the Figure~\ref{fig:hist}, it is also interesting to observe that the set of active items in factors (1 \& 2) vs factor 3 seem to be mostly disjoint.				
        
       This analysis provides a first attempt to analyze food shopping patterns for American consumers on a large scale. As the world, at large, struggles with a combination of increasing obesity rates and food insecurity, this analysis shows that consumer preferences are densely clustered across both stores and items. This indicates that household tend to have fairly rigid preferences over the stores where they shop. Furthermore, they tend to consume a relatively small number of products from the universe of available products. The concentration in both stores and products is indicative of limited search behavior and substantial behavioral rigidity which may be associated with suboptimal outcomes in terms of nutrition and health.
	
\vspace{-1em}
\subsection{Scalability}
\label{sec:scale}
\vspace{-0.3em}

\begin{figure}[!htbp]		
	\vskip -0.2in
	\begin{center}
		\centerline{\includegraphics[scale=0.29]{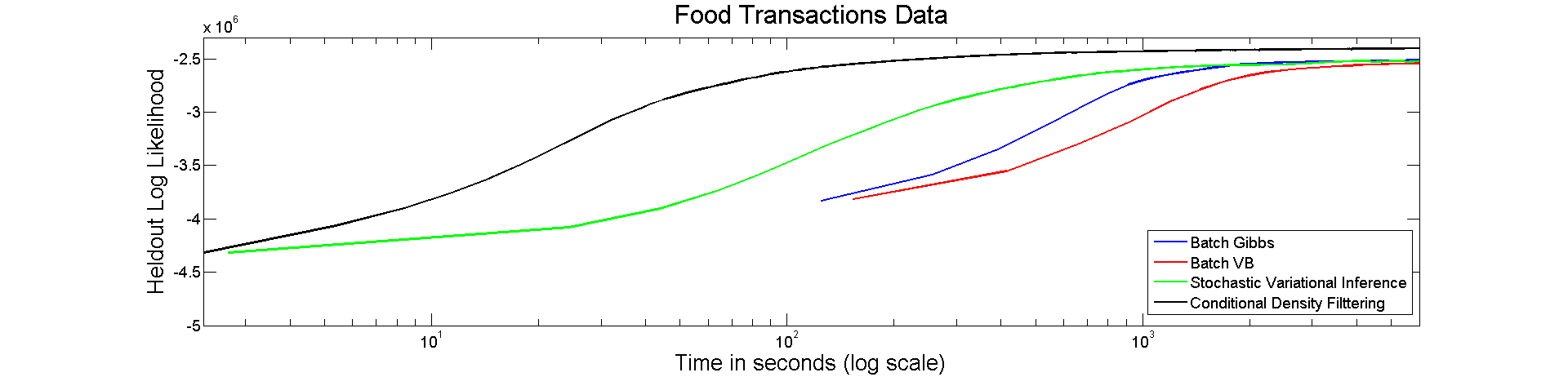}}
		\vskip -0.2in	
		\caption{Time vs heldout log likelihoods with various methods on transactions data}
		\label{fig:timecomparison}
	\end{center}
	\vskip -0.5in		
\end{figure}

We now perform an experiment comparing the proposed inference methods (batch and online) to assess their scalability (Figure~\ref{fig:timecomparison}). We first use the Transactions data ($117054\times 438 \times 67095$) for this experiment. We would like to note that the state-of-the-art methods for count-valued tensor, such as the Poisson Tensor Factorization (PTF) method from the Tensor Toolbox~\cite{chi2012tensors}, are simply infeasible to run on this data because of storage explosion issue (the method requires expensive flattening operations of the tensor). The other baseline \textsc{lraNTD}~\cite{lraNTD} we used in our experiments was also infeasible to run on this data. We set $R=100$ for each method (about 60 factors were found to be significant, based on the inferred values of the $\lambda_r$'s) and use a minibatch size of 100000 for all the online inference methods. For the conditional density filtering as well as stochastic variational inference, we set the learning rate as $t_0 = 0$ and $\kappa = 0.5$. Figure~\ref{fig:timecomparison} shows that online inference methods (conditional density filtering and stochastic variational inference) converge much faster to a good solution than batch methods. This experiment shows that our online inference methods can be computationally viable alternatives if their batch counterparts are slow/infeasible to run on such data. 

\begin{wrapfigure}{r}{0.5\textwidth} 
\vspace{-20pt}
  \begin{center}
    \includegraphics[scale=0.15]{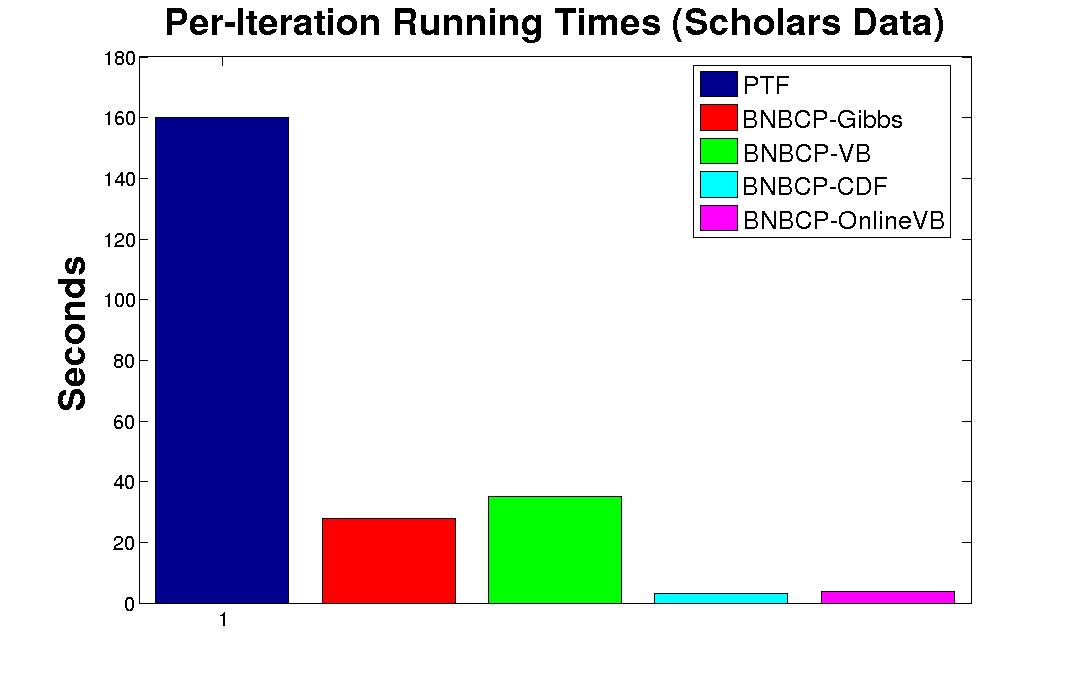}
				\caption{Timing comparison of various methods on Scholars data}
				\label{fig:dukecomp}
  \end{center}
  \vspace{-20pt}
  \vspace{1pt}
\end{wrapfigure}

We then perform another experiment on the Scholars data, on which the PTF method of~\cite{chi2012tensors} was feasible to run and compare its per-iteration running time with our model (using both batch as well as online inference). Since PTF cannot handle missing data, for this experiment, each method was run with all the data. As Fig~\ref{fig:dukecomp} shows, our methods have running times that are considerably smaller than that of PTF. 

\section{Conclusion}
\label{sec:concl}

We have presented a fully Bayesian framework for analyzing massive tensors with count data, and have designed a suite of scalable inference algorithms for handling massive tensor data. In addition to giving interpretable results and inferring the rank from the data, the proposed model can infer the distribution over objects in each of the tensor modes which can be useful for understanding groups of similar objects, and also for doing other types of qualitative analyses on such data, as shown by our various experiments on real-world data sets. Simplicity of the inference procedure also makes the proposed model amenable for parallel and distributed implementations. e.g., using MapReduce or Hadoop. The model can be a useful tool for analyzing data from diverse applications and scalability of the model opens door to the application of scalable Bayesian methods for analyzing massive multiway count data. 
\vspace{-1.5em}

\section*{Acknowledgments}

The research reported here was supported in part by ARO, DARPA, DOE, NGA and ONR. Any opinions, findings, recommendations, or conclusions are those of the
authors and do not necessarily reflect the views of the Economic Research
Service, U.S. Department of Agriculture. The analysis, findings, and
conclusions expressed in this paper also should not be attributed to
either Nielsen or Information Resources, Inc. (IRI). This research was conducted in collaboration with USDA under a Third
Party Agreement with IRI.

\bibliographystyle{acm}
\bibliography{bayesian_ntf}

\begin{thebibliography}{10}

\bibitem{tensoricassp}
{\sc Bazerque, J.~A., Mateos, G., and Giannakis, G.~B.}
\newblock Inference of poisson count processes using low-rank tensor data.
\newblock In {\em ICASSP\/} (2013).

\bibitem{beutel2014flexifact}
{\sc Beutel, A., Kumar, A., Papalexakis, E.~E., Talukdar, P.~P., Faloutsos, C.,
  and Xing, E.~P.}
\newblock Flexifact: Scalable flexible factorization of coupled tensors on
  hadoop.
\newblock In {\em SDM\/} (2014).

\bibitem{cappe2009line}
{\sc Capp{\'e}, O., and Moulines, E.}
\newblock On-line expectation--maximization algorithm for latent data models.
\newblock {\em Journal of the Royal Statistical Society: Series B (Statistical
  Methodology) 71}, 3 (2009), 593--613.

\bibitem{chi2012tensors}
{\sc Chi, E.~C., and Kolda, T.~G.}
\newblock On tensors, sparsity, and nonnegative factorizations.
\newblock {\em SIAM Journal on Matrix Analysis and Applications 33}, 4 (2012),
  1272--1299.

\bibitem{cichocki2009nonnegative}
{\sc Cichocki, A., Zdunek, R., Phan, A.~H., and Amari, S.}
\newblock {\em Nonnegative matrix and tensor factorizations: applications to
  exploratory multi-way data analysis and blind source separation}.
\newblock John Wiley \& Sons, 2009.

\bibitem{dunson2005bayesian}
{\sc Dunson, D.~B., and Herring, A.~H.}
\newblock Bayesian latent variable models for mixed discrete outcomes.
\newblock {\em Biostatistics 6}, 1 (2005), 11--25.

\bibitem{guhaniyogi2014bayesian}
{\sc Guhaniyogi, R., Qamar, S., and Dunson, D.~B.}
\newblock Bayesian conditional density filtering.
\newblock {\em arXiv preprint arXiv:1401.3632\/} (2014).

\bibitem{heinrich2009variational}
{\sc Heinrich, G., and Goesele, M.}
\newblock Variational bayes for generic topic models.
\newblock In {\em KI 2009: Advances in Artificial Intelligence}. Springer,
  2009, pp.~161--168.

\bibitem{hoffman2013stochastic}
{\sc Hoffman, M.~D., Blei, D.~M., Wang, C., and Paisley, J.}
\newblock Stochastic variational inference.
\newblock {\em The Journal of Machine Learning Research 14}, 1 (2013),
  1303--1347.

\bibitem{inah2015haten2}
{\sc Inah, J., Papalexakis, E.~E., Kang, U., and Faloutsos, C.}
\newblock Haten2: Billion-scale tensor decompositions.
\newblock In {\em ICDE\/} (2015).

\bibitem{johndrow2014tensor}
{\sc Johndrow, J.~E., Battacharya, A., and Dunson, D.~B.}
\newblock Tensor decompositions and sparse log-linear models.
\newblock {\em arXiv preprint arXiv:1404.0396\/} (2014).

\bibitem{jordan1999introduction}
{\sc Jordan, M.~I., Ghahramani, Z., Jaakkola, T.~S., and Saul, L.~K.}
\newblock An introduction to variational methods for graphical models.
\newblock {\em Machine learning 37}, 2 (1999), 183--233.

\bibitem{kang2012gigatensor}
{\sc Kang, U., Papalexakis, E., Harpale, A., and Faloutsos, C.}
\newblock Gigatensor: scaling tensor analysis up by 100 times-algorithms and
  discoveries.
\newblock In {\em KDD\/} (2012).

\bibitem{kolda2009tensor}
{\sc Kolda, T.~G., and Bader, B.~W.}
\newblock Tensor decompositions and applications.
\newblock {\em SIAM review 51}, 3 (2009), 455--500.

\bibitem{kozubowski2008distributional}
{\sc Kozubowski, T.~J., and Podg{\'o}rski, K.}
\newblock {\em Distributional properties of the negative binomial L{\'e}vy
  process}.
\newblock Mathematical Statistics, Centre for Mathematical Sciences, Faculty of
  Engineering, Lund University, 2008.

\bibitem{leetaru2013gdelt}
{\sc Leetaru, K., and Schrodt, P.~A.}
\newblock Gdelt: Global data on events, location, and tone, 1979--2012.
\newblock In {\em ISA Annual Convention\/} (2013), vol.~2, p.~4.

\bibitem{papalexakis2015parcube}
{\sc Papalexakis, E., Faloutsos, C., and Sidiropoulos, N.}
\newblock Parcube: Sparse parallelizable candecomp-parafac tensor
  decompositions.
\newblock {\em ACM Transactions on Knowledge Discovery from Data\/} (2015).

\bibitem{rai14tensor}
{\sc Rai, P., Wang, Y., Guo, S., Chen, G., Dunson, D., and Carin, L.}
\newblock Scalable bayesian low-rank decomposition of incomplete multiway
  tensors.
\newblock In {\em ICML\/} (2014).

\bibitem{schein2014}
{\sc Schein, A., Paisley, J., Blei, D.~M., and H, W.}
\newblock Inferring polyadic events with poisson tensor factorization.
\newblock In {\em NIPS Workshop\/} (2014).

\bibitem{schmidt2009probabilistic}
{\sc Schmidt, M., and Mohamed, S.}
\newblock Probabilistic non-negative tensor factorisation using markov chain
  monte carlo.
\newblock In {\em 17th European Signal Processing Conference\/} (2009).

\bibitem{shashua2005non}
{\sc Shashua, A., and Hazan, T.}
\newblock Non-negative tensor factorization with applications to statistics and
  computer vision.
\newblock In {\em ICML\/} (2005).

\bibitem{zhao2014bayesian}
{\sc Zhao, Q., Zhang, L., and Cichocki, A.}
\newblock Bayesian cp factorization of incomplete tensors with automatic rank
  determination.

\bibitem{lraNTD}
{\sc Zhou, G., Cichocki, A., and S., X.}
\newblock Fast nonnegative matrix/tensor factorization based on low-rank
  approximation.
\newblock {\em IEEE Transactions on Signal Processing 60}, 6 (2012),
  2928--2940.

\bibitem{zhou2012NBPFA}
{\sc Zhou, M., Hannah, L.~A., Dunson, D., and Carin, L.}
\newblock Beta-negative binomial process and poisson factor analysis.
\newblock In {\em AISTATS\/} (2012).

\end{thebibliography}

\end{document}